\documentclass[10pt,twocolumn,letterpaper]{article}

\usepackage{cvpr}              

\usepackage{amsfonts}
\usepackage{amsmath}
\usepackage{multirow}
\usepackage{booktabs}
\usepackage{makecell}
\usepackage{minted}
\usepackage{listings}
\usepackage{mathtools}
\definecolor{cvprblue}{rgb}{0.21,0.49,0.74}
\usepackage[pagebackref,breaklinks,colorlinks,allcolors=cvprblue]{hyperref}
\usepackage{graphicx}
\usepackage{makecell}
\usepackage{tikz}
\usetikzlibrary{calc,tikzmark}
\newcommand{\vlgap}{-4.5pt}

\newcommand{\vlbetween}[2]{%
  \begin{tikzpicture}[remember picture,overlay]
    \draw[line width=1pt]
      ($(#1.west)+(\vlgap,  9pt)$) -- 
      ($(#2.west)+(\vlgap, -16pt)$);   
  \end{tikzpicture}%
}

\def\paperID{*****} 
\def\confName{CVPR}
\def\confYear{2026}

\title{MEMO: Human-like Crisp Edge Detection Using Masked Edge Prediction}

\author{
Jiaxin Cheng$^{1}$ \qquad
Yue Wu$^{2}$ \qquad
Yicong Zhou$^{1}$\thanks{Corresponding author}\\[4pt]
$^{1}$Department of Computer and Information Science, University of Macau, Macau, China\\
$^{2}$Capital One, USA\\[4pt]
{\tt\small yc47434@um.edu.mo \qquad yue.wu@capitalone.com \qquad yicongzhou@um.edu.mo}
}

\begin{document}
\maketitle
\begin{abstract}
Learning-based edge detection models trained with cross-entropy loss often suffer from thick edge predictions, which deviate from the crisp, single-pixel annotations typically provided by humans. While previous approaches to achieving crisp edges have focused on designing specialized loss functions or modifying network architectures, we show that a carefully designed training and inference strategy alone is sufficient to achieve human-like edge quality. In this work, we introduce the Masked Edge Prediction MOdel (MEMO), which produces both accurate and crisp edges using only cross-entropy loss. We first construct a large-scale synthetic edge dataset to pre-train MEMO, enhancing its generalization ability. Subsequent fine-tuning on downstream datasets requires only a lightweight module comprising 1.2\% additional parameters. During training, MEMO learns to predict edges under varying ratios of input masking. A key insight guiding our inference is that thick edge predictions typically exhibit a confidence gradient: high in the center and lower toward the boundaries. Leveraging this, we propose a novel progressive prediction strategy that sequentially finalizes edge predictions in order of prediction confidence, resulting in thinner and more precise contours. Our method achieves visually appealing, post-processing-free, human-like edge maps and outperforms prior methods on crispness-aware evaluations. \href{https://github.com/cplusx/MEMO_Edge_Detection}{https://github.com/cplusx/MEMO\_Edge\_Detection}

\end{abstract}

\section{Introduction}

Recent years have witnessed the remarkable success of deep learning in addressing the edge detection problem~\cite{hed,rcf,bdcn,edter,diffedge,uaed,muge}. In this learning-based paradigm, edge detection is typically formulated as a binary classification task distinguishing edge pixels from the background. Accordingly, cross-entropy loss has become the default choice for optimizing such models.

However, models trained with cross-entropy loss often suffer from ambiguous predictions, manifested as thick edge regions rather than the crisp, one-pixel-wide contours typically provided by human annotators. To address this discrepancy and produce human-like sharp edges, several strategies have been proposed. These include adding sparsity-promoting losses focused on edge neighborhoods~\cite{ced,lpcb,cats}, using refined supervision labels~\cite{refined_label}, or recasting edge detection as a generative task powered by diffusion backbones~\cite{diffedge}. Despite these efforts, we observe that their crispness often falls below 50\% on benchmarks like BSDS~\cite{bsds} and Multicue~\cite{multicue}, where annotations come from multiple annotators and vary at the pixel level, introducing label ambiguity and softening the supervision signal.

In contrast to prior efforts that modify architectures or supervision, we argue that crisp edge prediction can be achieved through a carefully designed training and inference strategy alone. To this end, we propose the Masked Edge Prediction MOdel (MEMO), which incorporates a masked edge training scheme and a confidence-ordered inference mechanism. Our approach is motivated by a key observation: thick edge predictions typically display a confidence gradient, where central edge pixels exhibit the highest prediction confidence and peripheral regions gradually decay in certainty.

This observation inspires a simple yet effective strategy: finalize confident predictions first, then iteratively refine the uncertain ones. MEMO adopts a multi-step inference process to realize this idea. In the initial step, MEMO predicts the edge probabilities for all pixels in the image. Only a subset of highly confident predictions are retained, while the rest are deferred for further refinement. In subsequent steps, the model re-evaluates only the uncertain regions, producing new predictions and retaining the next most confident ones. This process continues iteratively, gradually completing the edge map from most to least certain predictions. 

However, enabling this behavior requires the model to learn from partially completed edge maps. This is because after each iteration, some pixels are finalized while the remaining ones are still masked. We therefore adopt masked edge training, where ground-truth edge pixels are randomly hidden at varying ratios and the model learns to recover them. Through this setup, MEMO learns to treat confident predictions as fixed and naturally suppress redundant activations in their neighborhoods, leading to sharper and more stable edge localization over iterations.

One challenge introduced by masked training is the increased risk of overfitting due to repeated exposure to the same small training set. To mitigate this, we construct a large-scale synthetic edge dataset using an instance segmentation model~\cite{sam} and pre-train MEMO on it. We then fine-tune the model on each real-world dataset using lightweight LoRA-based adapters~\cite{lora}, which improves generalization while keeping per-task training costs minimal.

Finally, recent works have shown growing interest in multi-granularity edge prediction~\cite{muge,sauge}, a task that aims to reveal edge structures at different levels of abstraction. We demonstrate that MEMO naturally supports this capability through classifier-free guidance~\cite{cfg}. This technique is originally introduced in diffusion models~\cite{ddpm} to control generation quality, and works by extrapolating between conditioned and unconditioned predictions, scaled by a user-defined guidance parameter. In our case, this parameter becomes a granularity scale and serves a different purpose, allowing MEMO to smoothly transition from sparse, high-level contours to dense, low-level details. Unlike previous methods~\cite{muge} that require supervision on paired granularity-edge annotations, MEMO achieves this multi-granularity behavior purely via inference-time adjustment, eliminating the need for additional labels or retraining.

Our main contributions are as follows:
\begin{itemize}
    \item We propose MEMO, a novel edge detection framework featuring masked edge training and confidence-ordered inference, capable of producing crisp, human-like edge. 
    \item We construct a large-scale synthetic edge dataset and demonstrate that pre-training MEMO on this dataset improves its generalization and downstream performance.
    \item We show that MEMO naturally supports multi-granularity edge prediction, enabling flexible control over edge density purely at inference time.
    \item Extensive experiments confirm that MEMO achieves state-of-the-art results under crispness-aware evaluation and exhibits superior visual alignment with human annotations, while maintaining strong performance under standard edge detection metrics.
\end{itemize}

\begin{figure*}[!t]
    \centering
    \includegraphics[width=\linewidth]{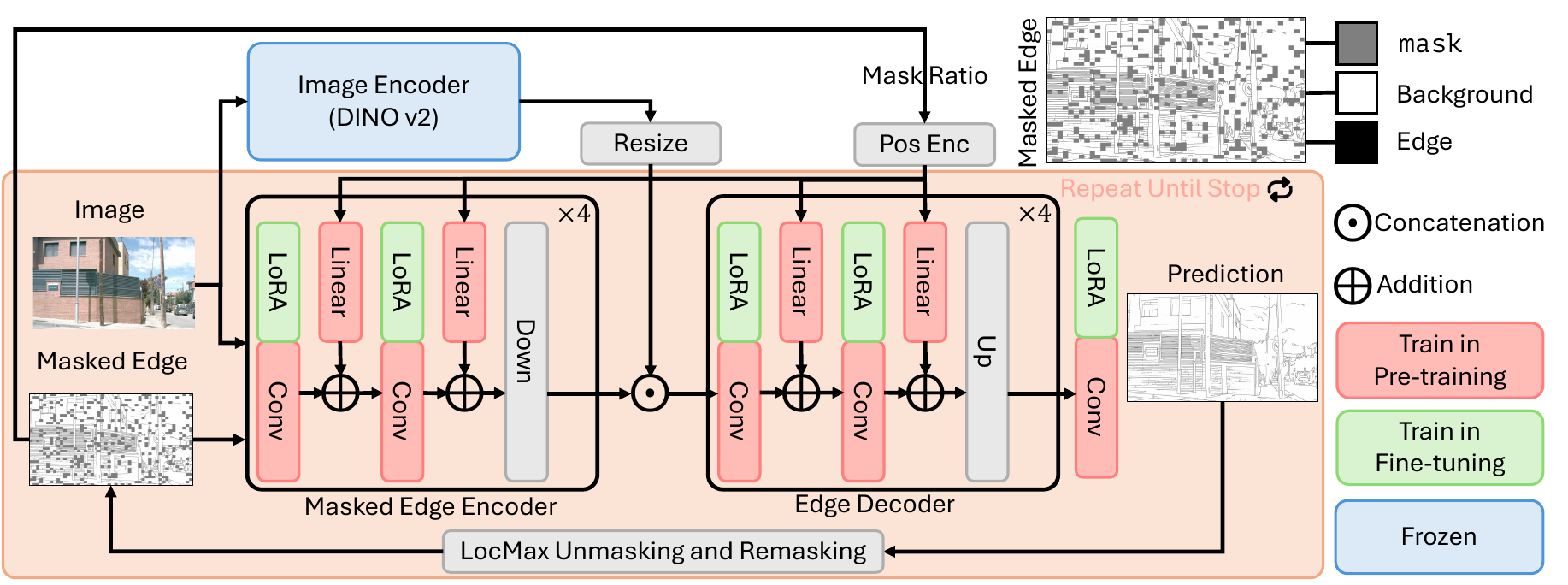}
    \caption{Overview of MEMO’s architecture. MEMO is trained in two stages. In the first stage, we train the masked edge encoder $F_E$ and edge decoder $D$ on a large-scale synthetic edge dataset while keeping the pre-trained image encoder $F_I$ frozen. In the second stage, we fine-tune on downstream datasets by injecting LoRA adapters into the edge encoder and decoder, while keeping all other components fixed. During inference, MEMO begins with a fully masked edge map and iteratively completes it based on a confidence-ordered strategy. At each step, the high confident predictions are finalized, and the remaining pixels are remasked for the next iteration according to the LocMax strategy in~\Cref{sec.conf_infer}. For better visualization, block-wise masks are illustrated, although masking is applied pixel-wise in practice.}
    \label{fig.pipeline}
\end{figure*}

\section{Method}

\subsection{Masked Edge Prediction Model (MEMO)}

Given an image $I \in \mathbb{R}^{h \times w}$, MEMO aims to predict a binary edge map $E \in \{0,1\}^{h \times w}$, where each pixel is classified as either edge or background. Unlike conventional approaches that perform edge prediction in a single forward pass, MEMO adopts a recursive inference process that progressively reveals the edge structure, as illustrated in~\Cref{fig.pipeline}. At the beginning of inference, the entire edge map is masked, with all pixels assigned a special \texttt{mask} symbol. In each iteration, MEMO predicts edge probabilities for the masked regions. A subset of high-confidence predictions is then finalized, while the rest are remasked for refinement in the next step. This iterative process continues until no further changes occur, as detailed in \Cref{sec.conf_infer}.

The model architecture consists of three components: an image encoder $F_I$, a masked edge encoder $F_E$, and a shared edge decoder $D$. Since the image $I$ remains static across iterations, it is encoded only once by $F_I$, while the changing masked edge map $E_r$ is re-encoded at every step. Following common practice~\cite{rcf,uaed,muge,pidinet}, we use a pre-trained vision backbone for $F_I$ and keep it frozen during training. The edge prediction is formulated as:
\begin{align}
    p(E \mid I, E_r) = \mathrm{Sigmoid}(D(F_I(I), F_E(I, E_r, r), r)),
\end{align}
where $E_r$ is the masked edge input at mask ratio $r \in (0,1]$.




\subsection{Masked Edge Training}
To equip MEMO with the ability to handle partially revealed edge maps during inference, we train it under varying levels of edge visibility using a masked edge training scheme. Specifically, for each training sample, we randomly select a masking ratio $r \in (0,1]$ and generate a masked edge map $E_r$ by applying independent Bernoulli masking at each pixel $i$:
\begin{equation}
  E_r[i] =
    \begin{cases}
      \texttt{mask} & \text{if } U(0,1) < r \\
      E[i] & \text{otherwise}
    \end{cases}
\end{equation}
where $U(0,1)$ is a uniform random variable.

To make the model aware of the masking ratio $r$, we embed it using a sinusoidal positional encoding~\cite{vaswani2017attention}, followed by a linear transformation. The resulting embedding is injected into features $f^l$ of every convolution block $l$ in the masked edge encoder $F_E$ and the decoder $D$:
\begin{equation}
    f^l = f^l + \mathrm{Linear}(\mathrm{PE}(r)) \quad \forall l \in \{F_E, D\}.
\end{equation}

The training objective is to reconstruct the masked entries of $E$ given $I$ and $E_r$. We apply a binary cross-entropy loss only on the masked pixels:
\begin{equation}
\begin{aligned}
\mathcal{L}
&= -\mathbb{E} \left[
\frac{1}{r} \sum_{i=1}^{h \times w} 
\mathbf{1}[E_r[i] = \texttt{mask}] \cdot \right. \\
&\quad \left(
E[i] \log p_i +
(1 - E[i]) \log(1 - p_i)
\right)
\Bigg],
\end{aligned}
\end{equation}
where $\mathbf{1}[\cdot]$ is the indicator function and $p_i \triangleq p(E[i] \mid I, E_r)$ is the predicted edge probability at pixel $i$.

\subsection{Confidence-Ordered Inference}\label{sec.conf_infer}

At each inference step, MEMO predicts edge probabilities for all currently masked pixels. Instead of finalizing all predictions at once, only a subset of high-confidence pixels are retained, while the rest are remasked and passed back into the model for further refinement.

A straightforward strategy would be to unmask a fixed percentage of pixels with the highest confidence scores. However, this naive approach can lead to overly thick edge regions, as high-confidence predictions often cluster spatially along strong edges. Consequently, selecting globally top-ranked pixels may finalize many adjacent predictions simultaneously, resulting in blurry or thick contours as shown in~\Cref{fig.qualitative_infe_strategy}.

To mitigate this issue, we propose a local maxima-based unmasking strategy, referred to as LocMax. Rather than relying on global confidence rankings, LocMax selects a pixel for finalization only if its confidence is the highest within its $3 \times 3$ local neighborhood. Formally, we use the confidence for each pixel $c_i = \max(p_i, 1 - p_i)$, and finalize its prediction according to:
\begin{equation}
  E_r[i] =
    \begin{cases}
      0 & \text{if } p_i < 0.5 \text{ and } c_i = \max(c_i \text{ in } 3 \times 3) \\
      1 & \text{if } p_i \geq 0.5 \text{ and } c_i = \max(c_i \text{ in } 3 \times 3) \\
      \mathtt{mask} & \text{otherwise}
    \end{cases}
\end{equation}

For non-masked pixels, the confidence is set to zero and thus does not affect local comparison. The LocMax strategy ensures convergence, as the number of masked pixels decreases monotonically after each iteration. However, the number of steps required to converge can vary across samples. We observe that although most pixels are resolved within the first 10--20 steps, full convergence may take up to 70 steps in some cases, which can be computationally expensive.

\begin{figure}[!h]
    \centering
    \includegraphics[width=\linewidth]{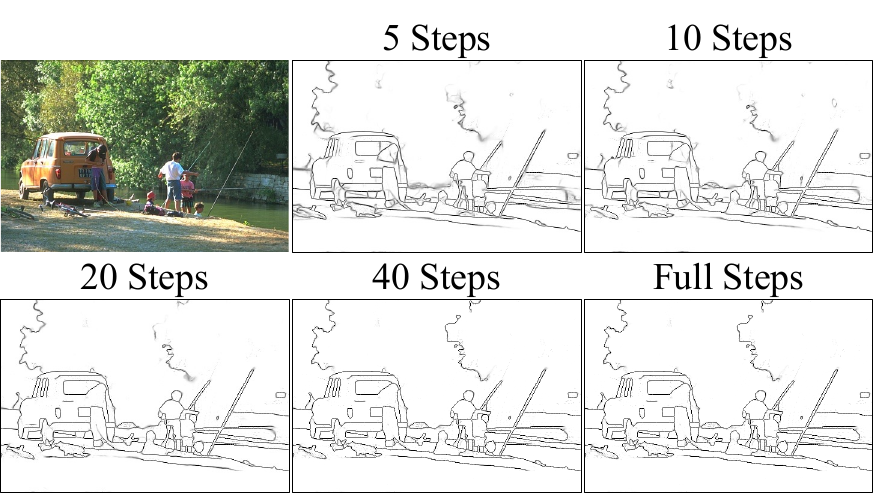}
    \caption{With very few steps, MEMO produces ambiguous edge maps. As the number of steps increases, the edges become sharper and more structurally coherent. Notably, using just 10 steps already yields visually crisp results.}
    \label{fig.qualitative_steps}
\end{figure}

To strike a balance between efficiency and accuracy, we adopt an early-stopping scheme. After a fixed number of steps, we stop remasking and retain all subsequent predictions. This significantly reduces inference time while maintaining visual quality. As illustrated in~\Cref{fig.qualitative_steps}, MEMO already produces crisp edges after 10 steps. Additional quantitative analysis is provided in~\Cref{sec.ablation}.

\begin{figure*}[t]
    \centering
    \includegraphics[width=\linewidth]{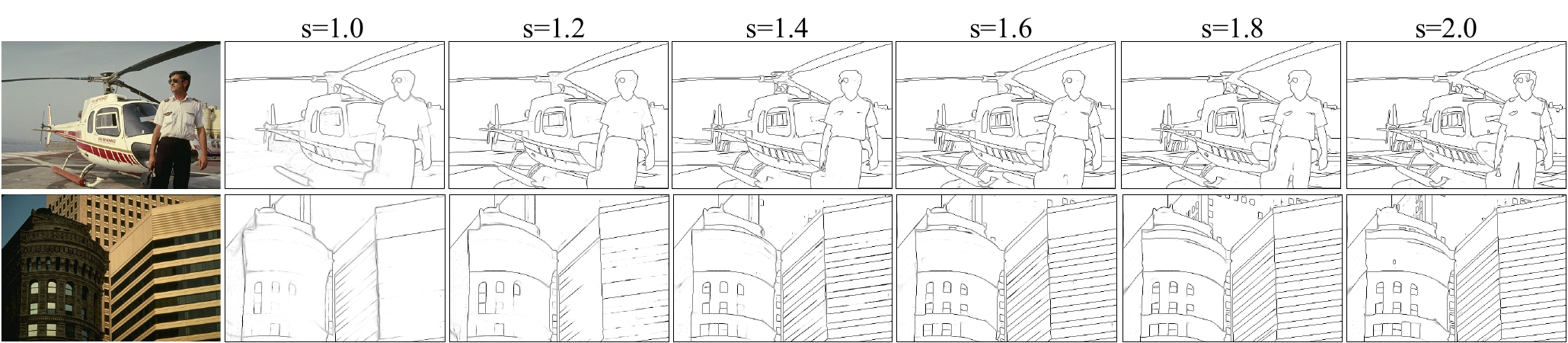}
    \caption{Multi-granularity prediction results of MEMO under varying granularity scales $s$. As the scale increases from $s=1.0$ to $s=2.0$, MEMO predicts increasingly dense edge maps with finer structural details. }
    \label{fig.qualitative_cfg}
\end{figure*}

\subsection{Multi-granularity Prediction}\label{sec.multi_granularity_pred}
Edge prediction is inherently ambiguous: different tasks may require varying levels of edge detail, and even for the same task, annotators may produce edge maps with significantly different densities. To accommodate such variability, it is desirable for an edge detection model to support multi-granularity prediction, \textit{i.e.}, the ability to generate edge maps of different levels of sparsity or detail~\cite{muge,sauge}.

MEMO enables this capability through a novel adaptation of classifier-free guidance~\cite{cfg}. In addition to standard training conditioned on both the image $I$ and masked edge $E_r$, we also train MEMO in an unconditioned setting, where the image is replaced with a zero-valued tensor $\varnothing$. This teaches MEMO to predict masked edges solely based on the visible structure in $E_r$, independent of image content.

During inference, we extrapolate between the image-conditioned and unconditioned predictions:
\begin{align}
    p(E \mid I, E_r) = \mathrm{Sigmoid}\big(s &\cdot D(F_I(I), F_E(I, E_r,r),r) \\\nonumber
    + (1-s) \cdot &D(F_I(\varnothing), F_E(\varnothing, E_r,r),r) \big)
\end{align}

Here, $s \geq 1$ is a user-defined granularity scale that controls the extrapolation. When $s=1$, the prediction is fully conditioned on the image and corresponds to standard inference. As $s$ increases, the model places more weight on the image-conditioned prediction, resulting in denser and crisper edge maps. This allows MEMO to smoothly transition from sparse to fine-grained edges. \Cref{fig.qualitative_cfg} illustrates this effect. As the granularity scale increases from $s=1.0$ to $s=2.0$, the predicted edge maps become progressively richer and more detailed. However, overly large values of $s$ may introduce excessive fine details and false positives, which can hurt evaluation precision. In practice, we find that scales in the range of $s=1.0$ to $s=2.0$ offer a good trade-off between edge density and accuracy. We further discuss extreme scale values in~\Cref{sec.extreme_granularity_scale}.

\subsection{Pre-training on Synthetic Dataset}

Under masked edge training, MEMO must learn to predict edges across a wide range of masking ratios. This increases the difficulty of optimization and requires the model to see more diverse training samples to avoid overfitting. However, existing edge datasets are typically small in scale, and repeatedly training on them can lead to poor generalization.

To address this, we pre-train MEMO on a large synthetic edge dataset before fine-tuning on real annotations. While collecting large-scale human-labeled edge maps is costly, synthetic edge maps can be obtained by extracting object boundary contours~\cite{rgbd}. We construct synthetic edge maps from images using the LAION dataset. Since LAION does not provide semantic boundaries, we use the Segment Anything Model (SAM)~\cite{sam} to automatically segment objects and generate instance masks. For each instance mask, we apply a morphological erosion and compute the difference between the original and eroded mask to obtain crisp binary contours. The final synthetic edge map for an image is attained by aggregating across all instances boundaries. In total, we collect 400,000 such image–edge pairs to pre-train MEMO. Visual examples of the synthetic dataset are provided in~\Cref{sec.appendix_dataset_example}

\section{Experiments}
\subsection{Experimental Settings}

\noindent\textbf{Dataset}  
We evaluate our model on three standard edge detection datasets. The BSDS~\cite{bsds} dataset contains 200 training, 100 validation, and 200 testing images of natural scenes. The BIPED~\cite{bipedv2} dataset focuses on street view imagery and includes 200 training and 50 testing images. The Multicue~\cite{multicue} dataset features outdoor scenes and comprises 80 training and 20 testing images.

\noindent\textbf{Metrics}  
Following~\cite{diffedge}, we report the Optimal Dataset Score (ODS) and Optimal Image Score (OIS), which are F1-scores computed after binarizing predictions at various thresholds and selecting the best threshold based on precision and recall of edge pixels. A predicted edge pixel is considered correct if it falls within 0.75\% of the image size. Both ODS and OIS are evaluated under the standard evaluation (SEval) and the crispness-aware evaluation (CEval). In SEval, predictions are post-processed with non-maximum suppression (NMS) and edge thinning. In CEval, raw predictions are evaluated without post-processing.

We assess edge quality and perceptual similarity to human annotations by using three metrics: Average Crispness~\cite{refined_label} (AC), Fréchet Inception Distance~\cite{fid} (FID), and Learned Perceptual Image Patch Similarity~\cite{lpips} (LPIPS). AC measures the proportion of edge pixels retained after applying NMS, with higher values indicating crisper predictions. FID and LPIPS evaluate distributional and perceptual similarity to ground-truth edges in feature space, with lower values indicating closer alignment to human annotations.


\noindent\textbf{Implementation Details}  
During pre-training, MEMO is trained using the AdamW optimizer~\cite{adamw} with a batch size of 64 and a learning rate of $5 \times 10^{-5}$. Input images are randomly cropped to $256 \times 256$ resolution. To preserve edge structures without aliasing, we apply minimal data augmentations, including just horizontal/vertical flips and 90-degree rotations. We use DINOv2-b~\cite{dinov2} as the frozen backbone image encoder. During fine-tuning on each dataset, we apply LoRA~\cite{lora} to train only adapter weights, keeping the pre-trained edge encoder and decoder frozen. Fine-tuning is conducted with a batch size of 32 and a learning rate of $2 \times 10^{-5}$. For BSDS, we use $256 \times 256$ resolution during training, while for BIPED and Multicue, we use $384 \times 384$. Evaluation is always performed on full-resolution images. To support multi-granularity prediction, we randomly drop 10\% of the image conditioning during training. \Cref{sec.appendix_network} provides the details of network architecture. 

\begin{figure*}
    \centering
    \includegraphics[width=\linewidth]{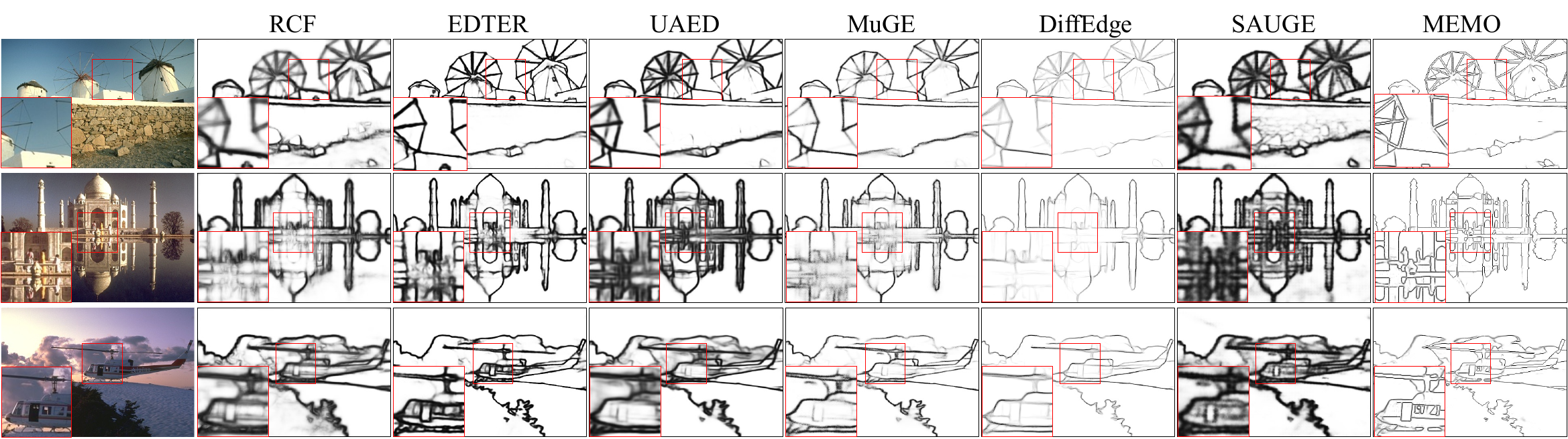}
    \caption{Qualitative comparison between MEMO and baseline methods on the BSDS dataset. MEMO produces crisp and structurally accurate edges, closely resembling human-style annotations. In contrast, baseline methods often produce thick or ambiguous edges, which increases the risk of merging closely spaced edges into a single contour. This leads to a loss of fine details and reduced separation between adjacent edges. MEMO avoids this by maintaining sharp and well-localized predictions, even in regions with densely packed edges.}
    \label{fig.qualitative_baselines}
\end{figure*}

\subsection{Binary Prediction Results}\label{sec.binary_results}
We evaluate MEMO on three benchmark datasets: BSDS, BIPED, and Multicue, with results reported in \Cref{tab.bsds_binary,tab.biped_binary,tab.multicue_binary}. Since MEMO supports multi-granularity prediction, its output varies with the granularity scale $s$ and the number of inference steps. For evaluation, we run MEMO using 11 values of $s$ ranging from 1.0 to 2.0 with interval 0.1, and report two results: C*, which corresponds to the scale that yields the best performance under the crispness-aware evaluation protocol (CEval), and AC*, which corresponds to the scale that achieves the highest AC. Each reported result uses the single best-performing scale for that metric. We use 10 inference steps on BSDS and Multicue, as this setting typically produces visually crisp predictions. For BIPED, we use 5 inference steps since high-quality edges can already be obtained with fewer iterations.

Quantitatively, MEMO achieves significant improvements under the CEval protocol and AC. These results demonstrate MEMO’s superior ability to produce crisp, human-like edge maps. Meanwhile, MEMO also retains competitive performance under the standard SEval protocol, consistently ranking among the top-performing methods. This suggests that enhancing crispness does not compromise standard edge detection accuracy.

Qualitative comparisons in \Cref{fig.qualitative_baselines} further support these findings. MEMO clearly outperforms all baselines by producing well-localized, non-ambiguous contours while preserving fine structures. In scenes with nearby edges or overlapping boundaries, baseline methods tend to either blur edges or produce irregular boundaries, whereas MEMO generates sharp and well-separated contours.

\begin{figure}[h]
    \centering
    \includegraphics[width=\linewidth]{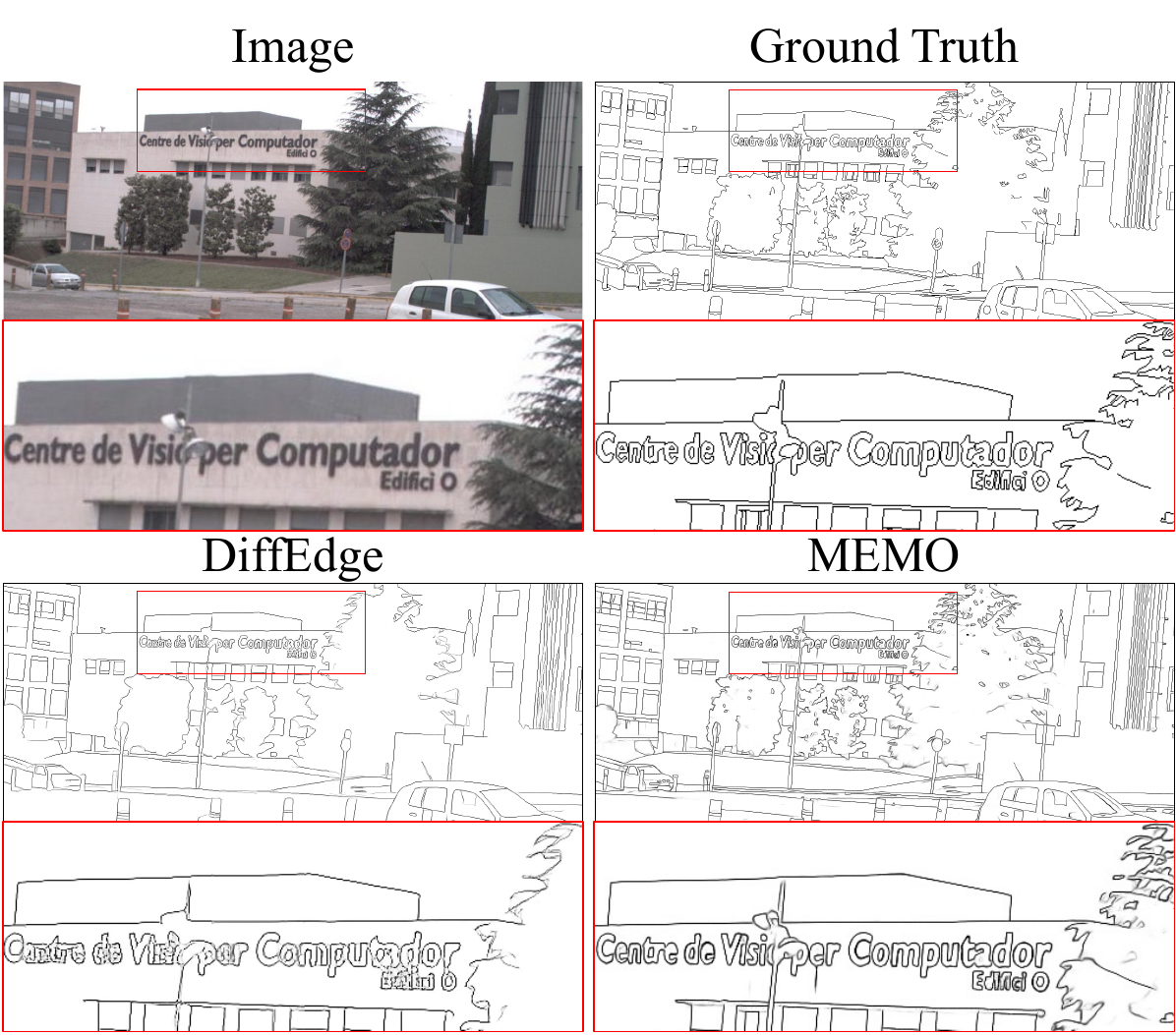}
    \caption{In regions containing dense and fine‑scale structures, such as text strokes and tree boundaries, DiffEdge~\cite{diffedge} exhibits inconsistent localization and produces fragmented or overly softened contours. In contrast, MEMO preserves high‑frequency edge details and yields coherent, well‑separated contours that more closely align with the ground‑truth annotations.}
    \label{fig.qualitative_biped}
\end{figure}

On BIPED, although DiffEdge~\cite{diffedge} achieves numerically strong scores, qualitative results reveal its limitations. As shown in \Cref{fig.qualitative_biped}, DiffEdge struggles to localize fine structures, often producing incomplete or visually unstable contours. In contrast, MEMO maintains high structural fidelity and produces edge maps that more closely resemble ground-truth annotations in both density and detail. These examples highlight MEMO’s robustness in cluttered scenes and reinforce its ability to capture edge structure with human-like precision.

\begin{table}[h]
    \caption{Standard and Crispness-aware evaluation on BSDS. All results are reported using a single-scale prediction. C*: Best CEval when $s=1.4$. AC*: Best AC when $s=1.8$.}
    \centering
    \setlength{\tabcolsep}{5pt}
    \begin{tabular}{lccccc}
    \toprule
    \multirow{2}{*}{\textbf{Methods}} & \multicolumn{2}{c}{\textbf{SEval}} & \multicolumn{2}{c}{\textbf{CEval}} & \multirow{2}{*}{\textbf{AC}}  \\
    \cmidrule(lr){2-3} \cmidrule(lr){4-5} 
    & ODS & OIS & ODS & OIS & \\
    \cmidrule(lr){1-1} \cmidrule(lr){2-2} \cmidrule(lr){3-3} \cmidrule(lr){4-4} \cmidrule(lr){5-5}  \cmidrule(lr){6-6}
    Canny~\cite{canny}   & 0.610 & 0.650 & - & - & - \\ 
    SE~\cite{se}      & 0.743 & 0.764 & - & - & - \\ 
    OEF~\cite{oef}     & 0.746 & 0.770 & - & - & - \\ 
    DeepCntr~\cite{deepcontour} & 0.757 & 0.776 & - & - & - \\ 
    DeepBdry~\cite{deepboundary}& 0.789 & 0.811 & - & - & - \\ 
    LPCB~\cite{lpcb}    & 0.800 & 0.816 & 0.693 & 0.700 & - \\
    CED~\cite{ced}     & 0.794 & 0.811 & 0.642 & 0.656 & 0.207 \\
    HED~\cite{hed}     & 0.788 & 0.808 & 0.588 & 0.608 & 0.215 \\ 
    RCF~\cite{rcf}     & 0.798 & 0.815 & 0.585 & 0.604 & 0.189 \\
    BDCN~\cite{bdcn}    & 0.806 & 0.826 & 0.636 & 0.650 & 0.233 \\
    PiDiNet~\cite{pidinet} & 0.789 & 0.803 & 0.578 & 0.587 & 0.202 \\
    EDTER~\cite{edter}   & 0.824 & 0.841 & 0.698 & 0.706 & 0.288 \\
    UAED~\cite{uaed}    & 0.829 & 0.847 & 0.722 & 0.731 & 0.227 \\
    MuGE~\cite{muge}    & 0.831 & 0.847 & 0.721 & 0.729 & 0.296\\
    DiffEdge~\cite{diffedge}& 0.834 & 0.848 & 0.749 & 0.754 & 0.476 \\
    SAUGE~\cite{sauge}   & \underline{0.847} & \textbf{0.868} & 0.687 & 0.699 & 0.197\\
    \cmidrule(lr){1-1} \cmidrule(lr){2-2} \cmidrule(lr){3-3} \cmidrule(lr){4-4} \cmidrule(lr){5-5}  \cmidrule(lr){6-6}
    MEMO (C*) & \textbf{0.854} & \underline{0.861} & \textbf{0.836} & \textbf{0.841} & \underline{0.663} \\ 
    MEMO (AC*) & 0.836 & 0.840 & \underline{0.820} & \underline{0.824} & \textbf{0.705} \\ 
    \bottomrule
    \end{tabular}
    \label{tab.bsds_binary}
\end{table}

\begin{table}[h]
    \caption{Standard and Crispness-aware evaluation on BIPED. C*: Best CEval when $s=1.7$. AC*: Best AC when $s=2.0$.}
    \centering
    \setlength{\tabcolsep}{5pt}
    \begin{tabular}{lccccc}
    \toprule
    \multirow{2}{*}{\textbf{Methods}} & \multicolumn{2}{c}{\textbf{SEval}} & \multicolumn{2}{c}{\textbf{CEval}} & \multirow{2}{*}{\textbf{AC}} \\
    \cmidrule(lr){2-3} \cmidrule(lr){4-5} 
    & ODS & OIS & ODS & OIS & \\
    \cmidrule(lr){1-1} \cmidrule(lr){2-2} \cmidrule(lr){3-3} \cmidrule(lr){4-4} \cmidrule(lr){5-5}  \cmidrule(lr){6-6}
    HED~\cite{hed}     & 0.829 & 0.847 & - & - & -\\ 
    RCF~\cite{rcf}     & 0.843 & 0.859 & 0.640 & 0.647 & 0.212\\
    BDCN~\cite{bdcn}    & 0.839 & 0.854 & 0.629 & 0.637 & 0.238\\
    DexiNed~\cite{dexined} & 0.859 & 0.867 & 0.764 & 0.776 & 0.295 \\
    DiffEdge~\cite{diffedge}& \textbf{0.899} & \textbf{0.901} & 0.856 & 0.863 & \underline{0.849} \\
    \cmidrule(lr){1-1} \cmidrule(lr){2-2} \cmidrule(lr){3-3} \cmidrule(lr){4-4} \cmidrule(lr){5-5}  \cmidrule(lr){6-6}
    MEMO (C*)   & \underline{0.888} & \underline{0.892} & \textbf{0.883} & \textbf{0.887} & 0.841\\ 
    MEMO (AC*)   & 0.878 & 0.881 & \underline{0.873} & \underline{0.877} & \textbf{0.858}\\ 
    \bottomrule
    \end{tabular}
    
    \label{tab.biped_binary}
\end{table}

\begin{table}[h]
    \caption{Standard and Crispness-aware evaluation on Multicue. C*: Best CEval when $s=1.4$. AC*: Best AC when $s=1.6$.}
    \centering
    \setlength{\tabcolsep}{5pt}
    \begin{tabular}{lccccc}
    \toprule
    \multirow{2}{*}{\textbf{Methods}} & \multicolumn{2}{c}{\textbf{SEval}} & \multicolumn{2}{c}{\textbf{CEval}} & \multirow{2}{*}{\textbf{AC}} \\
    \cmidrule(lr){2-3} \cmidrule(lr){4-5} 
    & ODS & OIS & ODS & OIS & \\
    \cmidrule(lr){1-1} \cmidrule(lr){2-2} \cmidrule(lr){3-3} \cmidrule(lr){4-4} \cmidrule(lr){5-5}  \cmidrule(lr){6-6}
    HED~\cite{hed}     & 0.851 & 0.864 & - & - & - \\ 
    RCF~\cite{rcf}     & 0.851 & 0.862 & 0.674 & 0.691 & 0.203 \\
    BDCN~\cite{bdcn}    & 0.891 & 0.898 & 0.676 & 0.679 & 0.249 \\
    DexiNed~\cite{dexined} & 0.872 & 0.881 & 0.764 & 0.776 & 0.274 \\
    EDTER~\cite{edter}   & \underline{0.894} & \underline{0.900} & 0.735 & 0.747 & 0.196 \\
    DiffEdge~\cite{diffedge}& \textbf{0.904} & \textbf{0.909} & 0.784 & 0.791 & 0.462 \\
    \cmidrule(lr){1-1} \cmidrule(lr){2-2} \cmidrule(lr){3-3} \cmidrule(lr){4-4} \cmidrule(lr){5-5}  \cmidrule(lr){6-6}
    MEMO (C*)  & \underline{0.894} & \underline{0.900} & \textbf{0.849} & \textbf{0.863} & \underline{0.626} \\ 
    MEMO (AC*)  & 0.885 & 0.892 & \underline{0.808} & \underline{0.835} & \textbf{0.659} \\ 
    \bottomrule
    \end{tabular}
    
    \label{tab.multicue_binary}
\end{table}

\noindent\textbf{Multi-Granularity Prediction}  
We further evaluate MEMO under the multi-granularity prediction setting. Following~\cite{muge}, each test image is evaluated under multiple granularity scales, and the highest performance across scales is aggregated per sample. For $M=11$, we use granularity scales from 1.0 to 2.0 with a step size of 0.1. For $M=3$, we use scales from 1.2 to 1.8 with a step size of 0.3. We compare MEMO against other multi-granularity-capable methods including UAED~\cite{uaed}, MuGE~\cite{muge}, and SAUGE~\cite{sauge}.

As shown in \Cref{tab.multigranularity}, MEMO achieves the best performance across all settings. This demonstrates MEMO’s strong capacity to adapt its predictions to varying annotation styles. Moreover, while some baselines perform well under SEval, they exhibit substantial performance drops under CEval, due to their reliance on post-processing for sharpening edges. MEMO, on the other hand, produces crisp predictions natively, enabling it to maintain high performance under CEval where no post-processing is allowed.

\begin{table}[h]
    \centering
    \caption{Multi-Granularity Evaluation on BSDS}
    \setlength{\tabcolsep}{5pt}
    \begin{tabular}{lllcccc}
    \toprule
    & & & UAED & MuGE & SAUGE & MEMO \\
    \cmidrule(lr){1-3} \cmidrule(lr){4-4} \cmidrule(lr){5-5} \cmidrule(lr){6-6} \cmidrule(lr){7-7}
    \multirow{4}{*}{\rotatebox[origin=c]{90}{\textbf{SEval}}} &
    \tikzmarknode{SEtop}{\multirow{2}{*}{\rotatebox[origin=c]{90}{M=3}}}
        & ODS & 0.838 & 0.845 & \underline{0.854} & \textbf{0.865} \\
       & & OIS & 0.847 & 0.854 & \underline{0.865} & \textbf{0.870} \\
    \cmidrule(lr){3-3} \cmidrule(lr){4-4} \cmidrule(lr){5-5} \cmidrule(lr){6-6} \cmidrule(lr){7-7}
    & 
    \tikzmarknode{SEbot}{\multirow{2}{*}{\rotatebox[origin=c]{90}{M=11}}}
        & ODS & 0.841 & 0.850 & \underline{0.857} & \textbf{0.873} \\
      &  & OIS & 0.847 & 0.856 & \underline{0.865} & \textbf{0.878} \\
    \midrule
    \multirow{4}{*}{\rotatebox[origin=c]{90}{\textbf{CEval}}} &
    \tikzmarknode{CEtop}{\multirow{2}{*}{\rotatebox[origin=c]{90}{M=3}}}
        & ODS & 0.739 & \underline{0.742} & 0.705 & \textbf{0.846} \\
       & & OIS & 0.746 & \underline{0.751} & 0.715 & \textbf{0.851} \\
    \cmidrule(lr){3-3} \cmidrule(lr){4-4} \cmidrule(lr){5-5} \cmidrule(lr){6-6} \cmidrule(lr){7-7}
    & 
    \tikzmarknode{CEbot}{\multirow{2}{*}{\rotatebox[origin=c]{90}{M=11}}}
        & ODS & \underline{0.747} & 0.745 & 0.706 & \textbf{0.855} \\
      & & OIS & \underline{0.756} & 0.752 & 0.715 & \textbf{0.860} \\
    \bottomrule
    \vlbetween{SEtop}{SEbot}
    \vlbetween{CEtop}{CEbot}
    \end{tabular}
    \label{tab.multigranularity}
\end{table}

\noindent\textbf{Visual Similarity to Human-Annotated Edges}  
A visually ideal edge map should be both crisp and perceptually aligned with human annotations, achieving high AC and low FID and LPIPS scores. We compare MEMO against top-performing edge detectors~\cite{muge,diffedge,sauge} and crispness-focused methods~\cite{cats,refined_label}. As shown in \Cref{tab.visual_similarity}, MEMO substantially improves AC and achieves the lowest FID and LPIPS, indicating that its predictions are not only sharper but also more visually faithful to human edges. Notably, while much prior work has assumed that additional supervision~\cite{refined_label} or specialized sparsity losses~\cite{cats,lpcb} are necessary to achieve human-like crispness, our results reveal that a well-designed training and inference strategy alone is sufficient to reach this quality level.

\begin{table}[h]
    \caption{Quantitative comparison of visual similarity to human annotations. MEMO outperforms both high-performance edge detectors~\cite{muge,diffedge,sauge} and crispness-focused methods~\cite{cats,refined_label}.}\label{tab.visual_similarity}
    \centering
    \setlength{\tabcolsep}{5pt}
    \begin{tabular}{lccc}
    \toprule
    \textbf{Methods}  & \textbf{AC} & \textbf{FID} & \textbf{LPIPS} \\
    \cmidrule(lr){1-1} \cmidrule(lr){2-2} \cmidrule(lr){3-3} \cmidrule(lr){4-4}
    MuGE~\cite{muge}    & 0.296 & 115.89 & 0.456\\
    DiffEdge~\cite{diffedge}& 0.476 & 89.96 & 0.300\\
    SAUGE~\cite{sauge}   & 0.197 & 243.01 & 0.651 \\
    \cmidrule(lr){1-1} \cmidrule(lr){2-2} \cmidrule(lr){3-3} \cmidrule(lr){4-4}
    CATS~\cite{cats} & 0.333 & 135.89 & 0.421 \\
    Refined Lbl~\cite{refined_label} & 0.424 & 135.29 & 0.370\\
    \cmidrule(lr){1-1} \cmidrule(lr){2-2} \cmidrule(lr){3-3} \cmidrule(lr){4-4}
    MEMO (C*) & \underline{0.663} & \underline{83.95} & \textbf{0.282}\\
    MEMO (AC*) & \textbf{0.705} & \textbf{75.55} & \underline{0.291} \\
    \bottomrule
    \end{tabular}
    
\end{table}

\subsection{Ablation Studies}\label{sec.ablation}

We conduct ablation studies on the BSDS dataset to examine the effects of different inference strategies, inference steps, and pre-training. Unless otherwise stated, we set the granularity scale to $s=1.4$, which achieves the highest crispness-aware evaluation score. Similar trends are observed when using the scale that maximizes AC.

\noindent\textbf{Effect of Inference Steps}  
\Cref{tab.inf_step} presents the performance of MEMO under different numbers of iterative inference steps. We evaluate five configurations: 5, 10, 20, 40, and a full-step setting without early-stopping.

As expected, increasing the number of steps generally improves both the CEval and crispness. However, this comes with a significant increase in inference time. Therefore, our remaining experiments adopts 10 inference steps as visual illustration in~\Cref{fig.qualitative_steps} suggests. Nevertheless, for applications that demand extremely sharp edges, MEMO can still benefit from more inference steps.

Interestingly, performance under the SEval protocol slightly decreases with more steps. We attribute this not to decreased accuracy, but rather to the nature of the evaluation. MEMO may predict edges that are visually valid yet absent from the ground-truth annotations due to incomplete labeling. These perceptually meaningful predictions are mistakenly counted as false positives, despite being correct from a human visual perspective. We discuss this phenomenon further in~\Cref{sec.discussion_more_inference}.

\begin{table}[h]
    \caption{MEMO's performance when using different inference unmasking steps. Inference time is measured on Nvidia 3090 GPU.}\label{tab.inf_step}
    \centering
    \small
    \setlength{\tabcolsep}{3pt}
    \begin{tabular}{llccccc}
        \toprule
        \multicolumn{2}{c}{Inf. Steps} & 5 & 10 & 20 & 40 & Full \\
        \cmidrule(lr){1-2} \cmidrule(lr){3-3} \cmidrule(lr){4-4} \cmidrule(lr){5-5}  \cmidrule(lr){6-6} \cmidrule(lr){7-7} 
        \multirow{2}{*}{\textbf{SEval}} & ODS & \textbf{0.855} & \underline{0.854} & 0.847 & 0.846 & 0.846 \\
         & OIS & \textbf{0.861} & \textbf{0.861} & 0.854 & 0.852 & 0.852 \\
        \cmidrule(lr){1-2} \cmidrule(lr){3-3} \cmidrule(lr){4-4} \cmidrule(lr){5-5}  \cmidrule(lr){6-6} \cmidrule(lr){7-7} 
        \multirow{2}{*}{\textbf{CEval}} & ODS & 0.835 & 0.836 & 0.838 & \underline{0.841} & \textbf{0.842}\\
         & OIS & 0.840 & 0.841 & 0.842 & \textbf{0.843} & \textbf{0.843} \\
        \cmidrule(lr){1-2} \cmidrule(lr){3-3} \cmidrule(lr){4-4} \cmidrule(lr){5-5}  \cmidrule(lr){6-6} \cmidrule(lr){7-7} 
        \multicolumn{2}{l}{\textbf{AC}} & 0.594 & 0.663 & 0.737 & \underline{0.828} & \textbf{0.840}\\
        \cmidrule(lr){1-2} \cmidrule(lr){3-3} \cmidrule(lr){4-4} \cmidrule(lr){5-5}  \cmidrule(lr){6-6} \cmidrule(lr){7-7} 
        \textbf{Inf. Time} & (sec.) & 0.62 & 1.33 & 2.77 & 5.63 & 10.46$\pm$1.03 \\
        \bottomrule
    \end{tabular}
\end{table}

\noindent\textbf{Inference Strategy}  
We further examine the impact of different unmasking strategies during inference. This experiment highlights that the choice of inference strategy is crucial for achieving both crisp and structurally accurate edge predictions. We compare our proposed LocMax strategy with two straightforward alternatives. Random unmasking: at each step, a random subset of masked pixels is finalized. TopK unmasking: at each step, the masked pixels with the highest confidence values are finalized, without considering their relative confidence compared to surrounding pixels. For all strategies, we perform 10 inference steps. For Random and TopK, we unmask an equal proportion of pixels at each step to ensure a comparable inference schedule.

\begin{figure}[h]
    \centering
    \includegraphics[width=\linewidth]{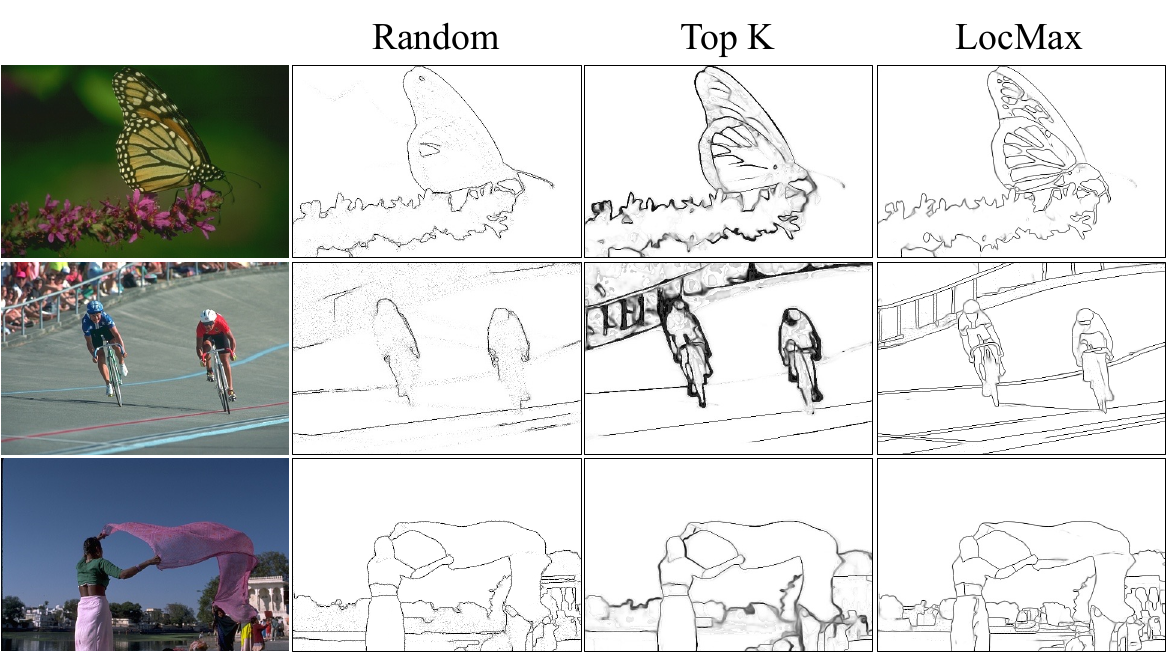}
    \caption{Qualitative comparison of MEMO when using different unmasking strategy. The random unmasking produces low quality prediction with discontinued edges. The TopK unmasking can still cause concentrated edges. The proposed LocMax unmasking scheme yields crisp and good coherent edge predictions. }
    \label{fig.qualitative_infe_strategy}
\end{figure}

\Cref{fig.qualitative_infe_strategy} shows qualitative comparisons. Random unmasking yields highly fragmented edge maps, where many contours break apart due to inconsistent local decisions. TopK, while more stable, often produces thick edge clusters as spatially correlated high-confidence pixels are spatially concentrated and are finalized simultaneously. In contrast, the proposed LocMax strategy finalizes a pixel only when it has the highest confidence in its local neighborhood, effectively preserving thin boundaries while maintaining contour continuity. As a result, LocMax produces edge maps that are both crisp and structurally coherent.

The quantitative results in \Cref{tab.inf_strategy} corroborate this observation. Random unmasking achieves high AC but suffers from poor detection accuracy. TopK performs better under SEval due to post-processing but exhibits large degradation under CEval. Only LocMax consistently performs well across all metrics, confirming that local confidence ordering is key to stable and high-quality edge prediction.

\begin{table}[h]
    \caption{MEMO's performance using different inference strategies}\label{tab.inf_strategy}
    \centering
    \setlength{\tabcolsep}{5pt}
    \begin{tabular}{llccc}
        \toprule
        \multicolumn{2}{c}{Inference Strategy} & Random & TopK & LocMax  \\
        \cmidrule(lr){1-2} \cmidrule(lr){3-3} \cmidrule(lr){4-4} \cmidrule(lr){5-5} 
        \multirow{2}{*}{\textbf{SEval.}} & ODS & 0.819 & \underline{0.825} & \textbf{0.854} \\
         & OIS & 0.829 & \underline{0.843} & \textbf{0.861} \\
        \cmidrule(lr){1-2} \cmidrule(lr){3-3} \cmidrule(lr){4-4} \cmidrule(lr){5-5} 
        \multirow{2}{*}{\textbf{CEval.}} & ODS & \underline{0.794} & 0.715 & \textbf{0.836} \\
         & OIS & \underline{0.804} & 0.720 & \textbf{0.841}  \\
        \cmidrule(lr){1-2} \cmidrule(lr){3-3} \cmidrule(lr){4-4} \cmidrule(lr){5-5} 
        \multicolumn{2}{l}{\textbf{AC}} & \textbf{0.671} & 0.510 & \underline{0.663} \\
        \bottomrule
    \end{tabular}
\end{table}

\noindent\textbf{Effect of Pre-training} We compare three training settings for MEMO: trained only on synthetic data, only on real data, and pre-trained on synthetic followed by fine-tuning on real data. \Cref{tab.pre_train_effect} reports results on BSDS under the same multi-granularity inference configuration. The synthetic-only model achieves the highest average crispness, as the synthetic dataset provides consistent, single-edge annotations without ambiguity. Fine-tuning on BSDS slightly reduces crispness due to the multi-annotator nature of BSDS, where multiple valid edge interpretations lead to local label inconsistencies. However, detection performance for both SEval and CEval are improved. This demonstrates that synthetic pre-training enhances generalization while largely preserving crisp edge quality.

\begin{figure}
    \centering
    \includegraphics[width=\linewidth]{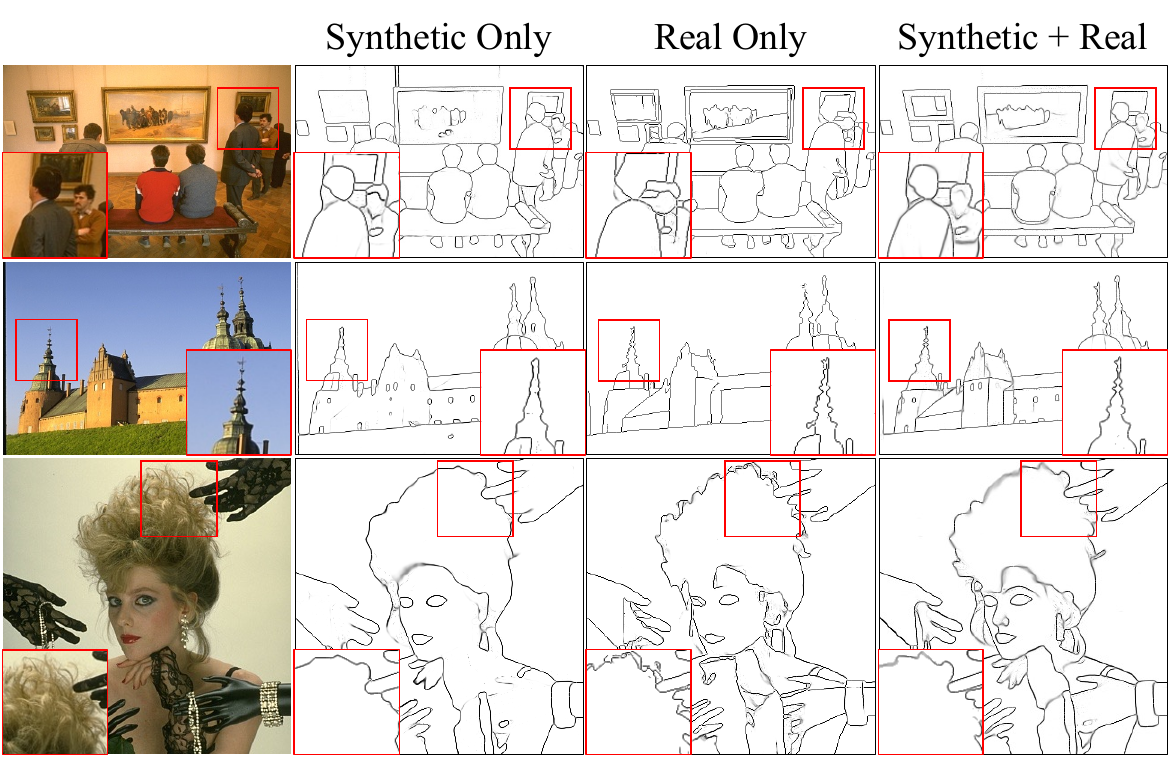}
    \caption{When trained only on real data, MEMO can produce duplicated or echoed contours in regions with ambiguous structure. Pre-training on synthetic data suppresses these multi-line artifacts and yields clean, single-edge predictions.}
    \label{fig.qualitative_training_set}
\end{figure}

In contrast, when MEMO is trained only on the real dataset, the predicted edges often exhibit edge duplication, where closely neighboring contours are activated simultaneously in regions with ambiguous structure, such as hair strands and object boundaries. As shown in \Cref{fig.qualitative_training_set}, this results in subtle multi-line artifacts rather than a single clean contour. Pre-training on synthetic data provides a strong bias toward producing a single, well-localized edge, effectively preventing such duplication and yielding cleaner and more consistent predictions.



\begin{table}[h]
    \caption{Performance of MEMO when trained on synthetic-only, real-only, and both dataset.}\label{tab.pre_train_effect}
    \centering
    \setlength{\tabcolsep}{5pt}
    \begin{tabular}{llccc}
        \toprule
        \multicolumn{2}{c}{Training Set} & Synth. & Real & Both  \\
        \cmidrule(lr){1-2} \cmidrule(lr){3-3} \cmidrule(lr){4-4} \cmidrule(lr){5-5} 
        \multirow{2}{*}{\textbf{SEval.}} & ODS & 0.729 & \underline{0.804} & \textbf{0.854} \\
         & OIS & 0.748 & \underline{0.810} &  \textbf{0.861} \\
        \cmidrule(lr){1-2} \cmidrule(lr){3-3} \cmidrule(lr){4-4} \cmidrule(lr){5-5} 
        \multirow{2}{*}{\textbf{CEval.}} & ODS & 0.725 & \underline{0.799} & \textbf{0.836} \\
         & OIS & 0.726 & \underline{0.803} &  \textbf{0.841} \\
        \cmidrule(lr){1-2} \cmidrule(lr){3-3} \cmidrule(lr){4-4} \cmidrule(lr){5-5} 
        \multicolumn{2}{l}{\textbf{AC}} & \textbf{0.765} & 0.619 & \underline{0.663} \\
        \bottomrule
    \end{tabular}
\end{table}

\section{Conclusion}
In this work, we present MEMO for human-like crisp edge detection. MEMO introduces a masked edge prediction training strategy and a confidence-ordered inference scheme, enabling crisp and well-localized edge maps without requiring specialized losses or complex model architectures. Moreover, MEMO naturally supports multi-granularity prediction through an inference-time extrapolation between conditioned and unconditioned predictions. With pre-training on our newly constructed synthetic edge dataset, MEMO further improves generalization on downstream datasets. Extensive quantitative and qualitative results demonstrate that MEMO achieves state-of-the-art performance in both crispness-aware evaluation and perceptual similarity to human annotations, while maintaining competitive accuracy under the standard evaluation protocol.


\section{Acknowledgment}
This work was funded in part by the Science and Technology Development Fund, Macau SAR (File no. 0050/2024/AGJ), by the University of Macau and University of Macau Development Foundation (File no. MYRG-GRG2024-00181-FST-UMDF).
{
    \small
    \bibliographystyle{ieeenat_fullname}
    \bibliography{main}
}

\clearpage
\appendix
\setcounter{page}{1}
\maketitlesupplementary



\section{Network Details}
\label{sec.appendix_network}
The masked edge encoder and edge decoder have a symmetric architecture composed of four residual blocks~\cite{resnet} with 128, 256, 512, and 768 output channels, respectively. Each residual block consists of two convolutional sub-blocks. In each convolutional sub-block, the input feature is processed by a group normalization layer~\cite{wu2018group}, a SiLU activation~\cite{silu}, and a $3\times3$ convolutional layer.

Overall, MEMO contains 238M parameters during pre-training, of which 86M belong to the DINOv2~\cite{dinov2} image encoder. After inserting the LoRA adapters~\cite{lora}, the total number of parameters increases to 241M, corresponding to only a 1.2\% increase.

\section{Flexibility of Granularity Scale}\label{sec.extreme_granularity_scale}

In \Cref{sec.multi_granularity_pred}, we discussed how adjusting the granularity scale controls the richness of predicted edges to support multi-granularity edge prediction. Based on visual observations, we typically set the scale in the range of 1.0 to 2.0. However, MEMO's prediction mechanism supports a much wider range of scale values, offering greater flexibility than prior methods.

\begin{figure}[!h]
    \centering
    \includegraphics[width=\linewidth]{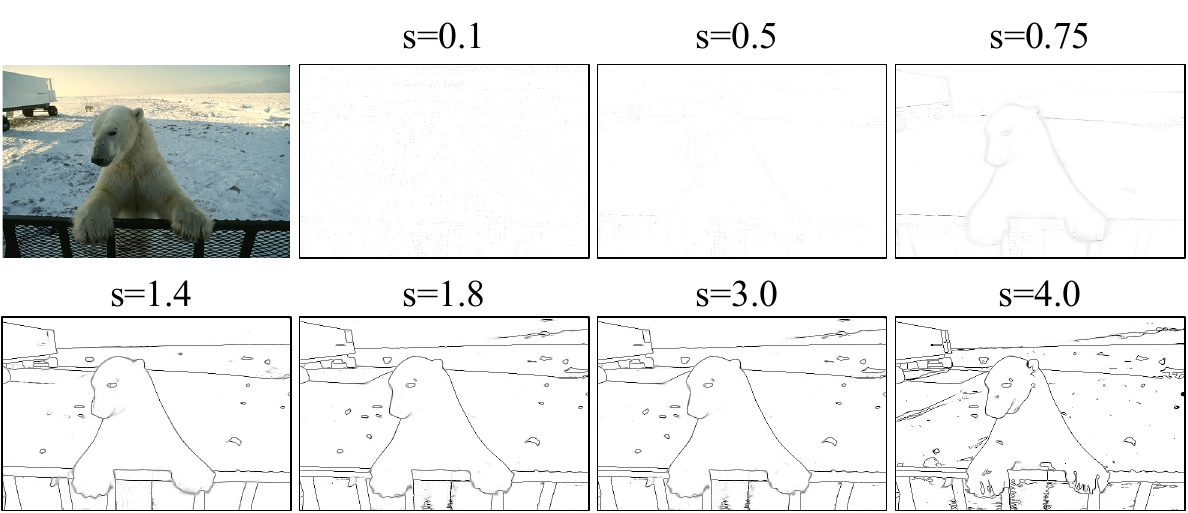}
    \caption{Edge predictions under varying granularity scale values. As the scale increases, more edges are progressively revealed. MEMO maintains high edge quality across a wide range of scales, but excessively large values may introduce spurious edges. This demonstrates MEMO’s flexible granularity control beyond pre-defined ranges used in prior work.}
    \label{fig.extreme_cfg}
\end{figure}

Previous work, such as MuGE~\cite{muge} and SAUGE~\cite{sauge}, enables granularity control using either paired supervision or multi-layer interpolation. However, both are limited to a pre-defined and fixed control range. In contrast, MEMO supports any positive scale value. As shown in \Cref{fig.extreme_cfg}, setting the scale close to zero suppresses all edges, while increasing the scale progressively reveals more edges. When the scale is moderately increased (\textit{e.g.}, $s=1.4$ achieves C* or $s=1.8$ achieves AC*), MEMO produces clean and complete edges. Pushing the scale further (\textit{e.g.}, $s=4.0$) continues to increase edge richness, but at the cost of more false positives and spurious structures. This illustrates that although MEMO is highly flexible, extreme values should be used cautiously depending on the application needs.

\section{Additional Discussion on Inference Steps}
\label{sec.discussion_more_inference}

\begin{figure}[!h]
    \centering
    \includegraphics[width=\linewidth]{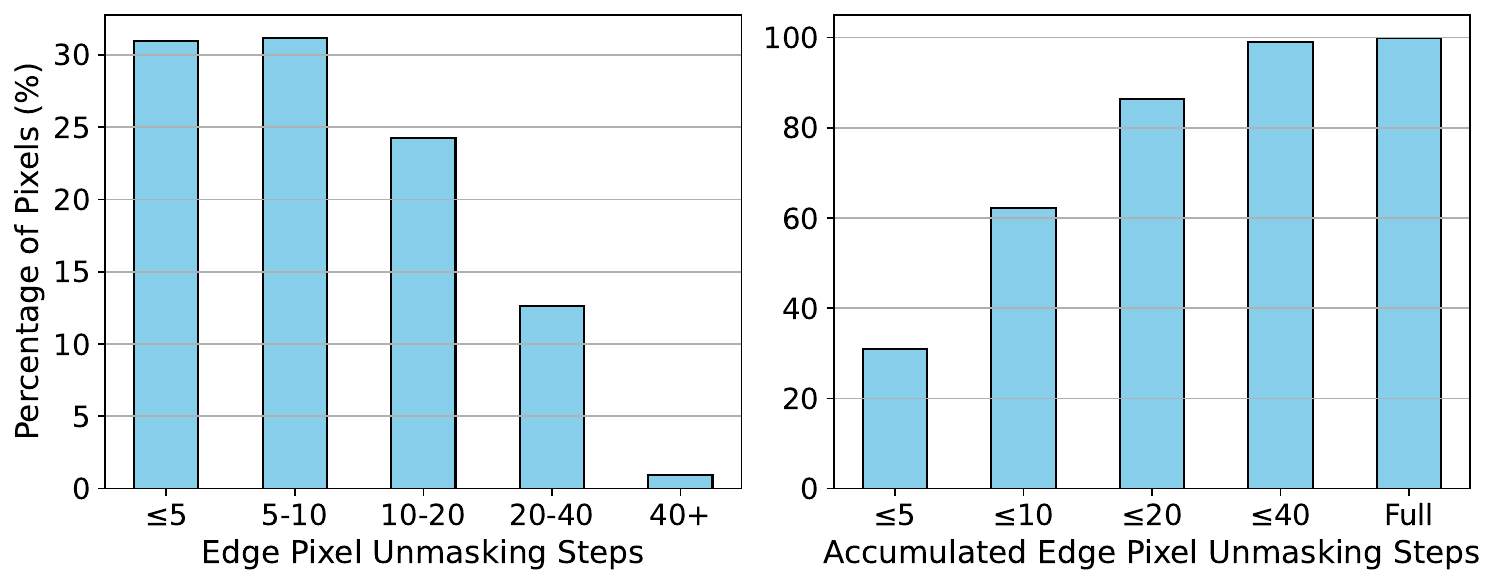}
    \caption{Distribution of edge pixel finalization steps across the inference process. Left: Histogram showing the percentage of edge pixels finalized at different unmasking step ranges. Right: Cumulative proportion of finalized pixels by step count.}
    \label{fig.edge_pixel_unmasking_distribution}
\end{figure}

\begin{figure}[!h]
    \centering
    \includegraphics[width=\linewidth]{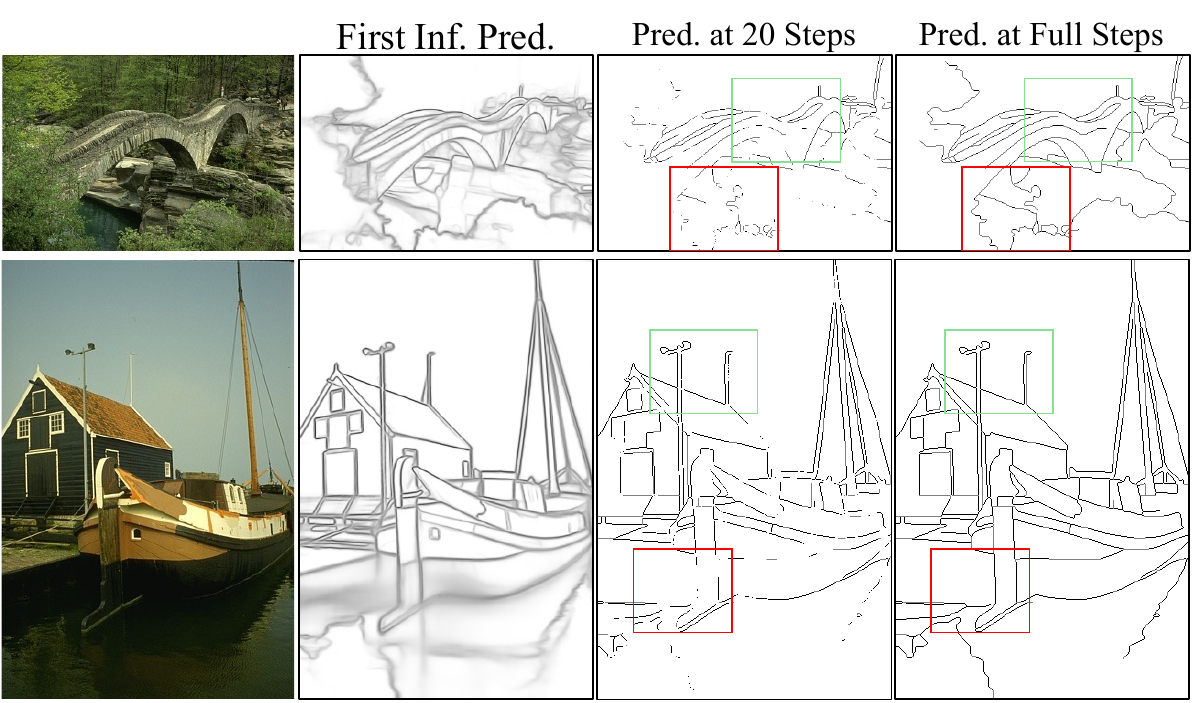}
    \caption{Intermediate predictions at different inference stages of MEMO. The second column shows edge predictions after the first inference pass, where all pixels are visualized. These predictions, similar to conventional methods, reveal thick and ambiguous prediction. However, the edges of different ambiguity converge at different speed. The last two columns show the finalized edge pixels after 20 steps and full inference, respectively. Less ambiguous boundaries usually converge early (green boxes), while highly uncertain regions (red boxes) require more iterations to stabilize. }
    \label{fig.intermediate_inference}
\end{figure}

While \Cref{sec.multi_granularity_pred} notes that MEMO typically finalizes most predictions within the first 10–20 steps, a trend also validated in the unmasking distribution shown in \Cref{fig.edge_pixel_unmasking_distribution}, we find that the actual convergence speed is sample‑dependent and strongly influenced by edge ambiguity. In particular, ambiguous edges that initially appear thick, fuzzy, or spatially uncertain tend to require more iterations to resolve. This behavior is illustrated in \Cref{fig.intermediate_inference}. The second column shows MEMO’s prediction after the first inference pass, which resembles outputs from existing methods: many edges are still thick and blurry. Yet these ambiguities are not uniform, some boundaries already appear relatively well-formed, while others remain diffuse and unstable.

As inference progresses, these edges converge at different speeds, as shown in the last two columns of \Cref{fig.intermediate_inference}. Clear and confident boundaries typically stabilize within 20 steps  as shown in green boxes, whereas ambiguous structures require substantially more iterations before reaching a consistent, thin contour as demonstrated in red boxes.

This observation suggests that the ideal number of inference steps should adapt to the visual complexity of the scene. It also explains why, as mentioned in \Cref{sec.binary_results}, BIPED achieves visually crisp results with 5 steps, while it takes more steps for BSDS and Multicue to reveal sufficient crisp edges.

\begin{figure}[!h]
    \centering
    \includegraphics[width=\linewidth]{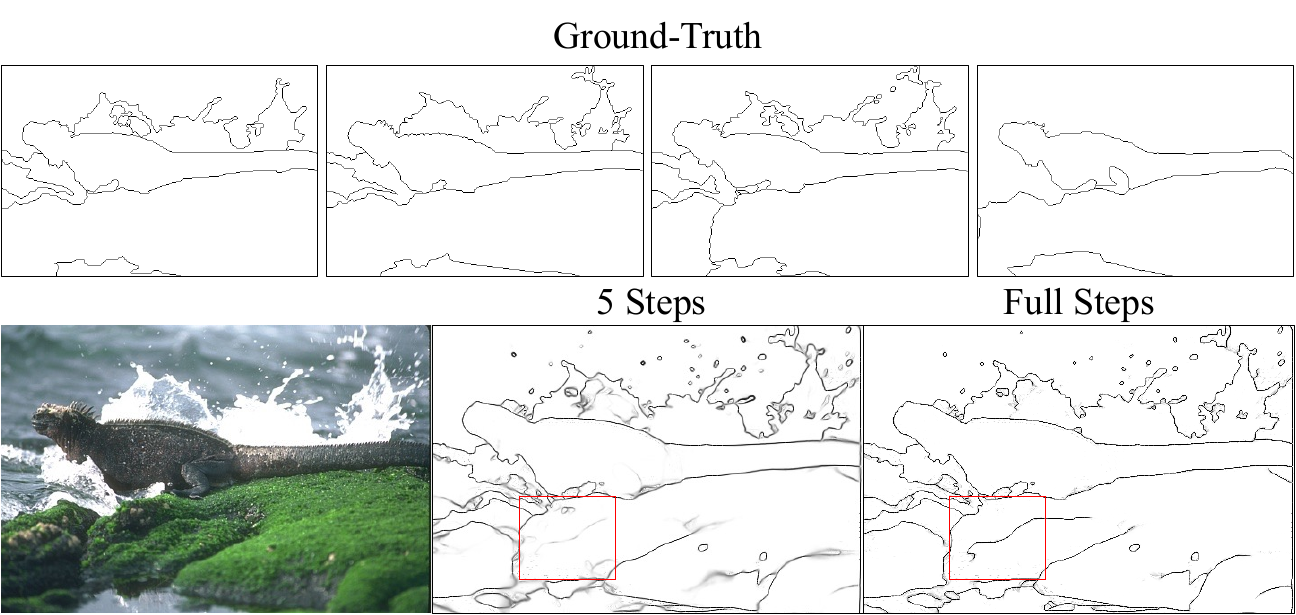}
    \caption{With fewer steps (5 steps), ambiguous edges are partially predicted with low confidence, resulting in faint contours. With full-step inference, MEMO finalizes these edges, leading to sharper but potentially unmatched predictions due to missing ground-truth labels. Although evaluation protocol may penalize these as false positives, they remain visually plausible from a human perspective.}
    \label{fig.inference_step_discussion}
\end{figure}

In~\Cref{sec.ablation}, we observe that MEMO's performance under the SEval protocol slightly declines as the number of inference steps increases. However, this is not due to reduced prediction accuracy, but rather stems from the limitations of the evaluation metric. As illustrated in \Cref{fig.inference_step_discussion}, edges that are ambiguous typically receive lower confidence scores in early inference steps. As a result, when using a small number of steps (\textit{e.g.}, 5 steps), these regions are often not finalized, leading to faint or partial predictions as shown in the red box. However, with more inference steps, MEMO is more likely to finalize these edges, especially if parts of them are progressively resolved with sufficient confidence. This behavior can cause perceptually valid but unannotated edges to be fully predicted, which are then incorrectly penalized as false positives. Despite the slightly lower SEval score at full-step inference, we argue that these predictions still reflect human visual perception and indicate strong prediction quality.

\section{MEMO vs Edge Post-process}\label{sec.appendix_memo_vs_post_process}


A common approach in edge detection pipelines is to rely on post-processing techniques, such as edge non-maximum suppression (NMS) or thinning to improve visual sharpness. These steps are often sufficient to boost benchmark scores under evaluation protocols, which permit a small spatial tolerance between predictions and ground-truth labels. However, high numerical scores do not necessarily imply high perceptual quality. In fact, these post-processed edges can diverge significantly from human-like annotations.

\begin{figure}[h]
    \centering
    \includegraphics[width=\linewidth]{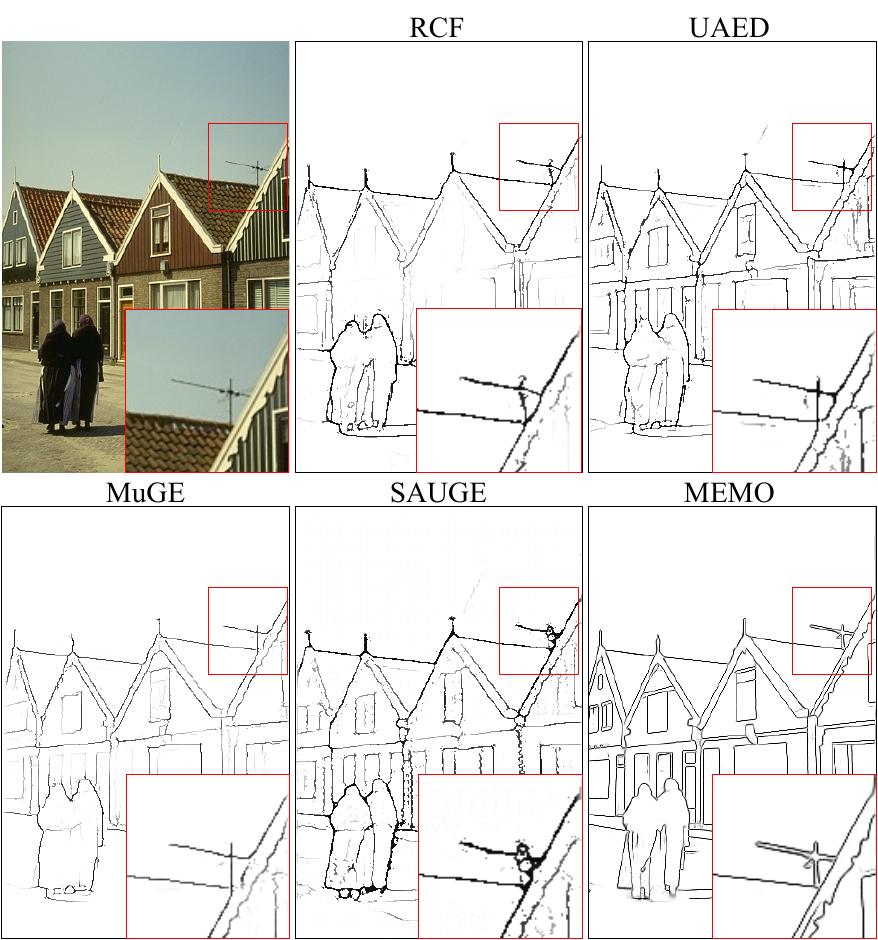}
    \caption{Qualitative comparison between MEMO without post-processing and baseline methods with post-processing. Although post-processing enhances benchmark scores, it does not guarantee perceptually faithful edge maps. Baseline predictions often exhibit duplicated, broken, or jittery edges, particularly in fine structures like the antenna and rooflines, due to the inherent ambiguity in thick or overlapping predictions. In contrast, MEMO directly predicts crisp and coherent edge structures, accurately capturing both global layouts and fine details without the need for post-processing.}
    \label{fig.compare_to_nms}
\end{figure}

As shown in \Cref{fig.compare_to_nms}, baseline methods~\cite{rcf,uaed,muge,sauge} that heavily depends on post-processing can produce spatially unstable edge predictions that are only superficially refined by thinning. For example, the antenna on the rooftop and the contours of the buildings frequently appear jagged, duplicated, or spatially misaligned even after post-processing. Such artifacts are tolerated by the evaluation metric but appear unnatural and imprecise to human observers. This becomes a critical limitation in applications where fine structure integrity or precise boundary localization is important.

MEMO addresses this issue from the root by incorporating crispness awareness directly into its prediction process. Rather than depending on external post-processing to refine ambiguous or overlapping predictions, MEMO progressively finalizes confident pixels while learning to suppress redundant activations in high-density regions. As a result, MEMO generates structurally coherent, well-localized, and perceptually aligned edge maps without any post-processing. This end-to-end design not only avoids the brittleness of hand-crafted post-processing rules but also leads to better generalization across different edge styles and levels of detail.

\section{Example of Synthetic Datasets}\label{sec.appendix_dataset_example}
When constructing the synthetic dataset, we follow the hyper‑parameter settings recommended by SAM~\cite{sam}, with one modification: we set \texttt{stability\_score\_thresh} to 0.85. This increases the number of recalled objects and, in turn, provides a richer set of edge candidates for training.

\begin{figure*}
    \centering
    \includegraphics[width=\linewidth]{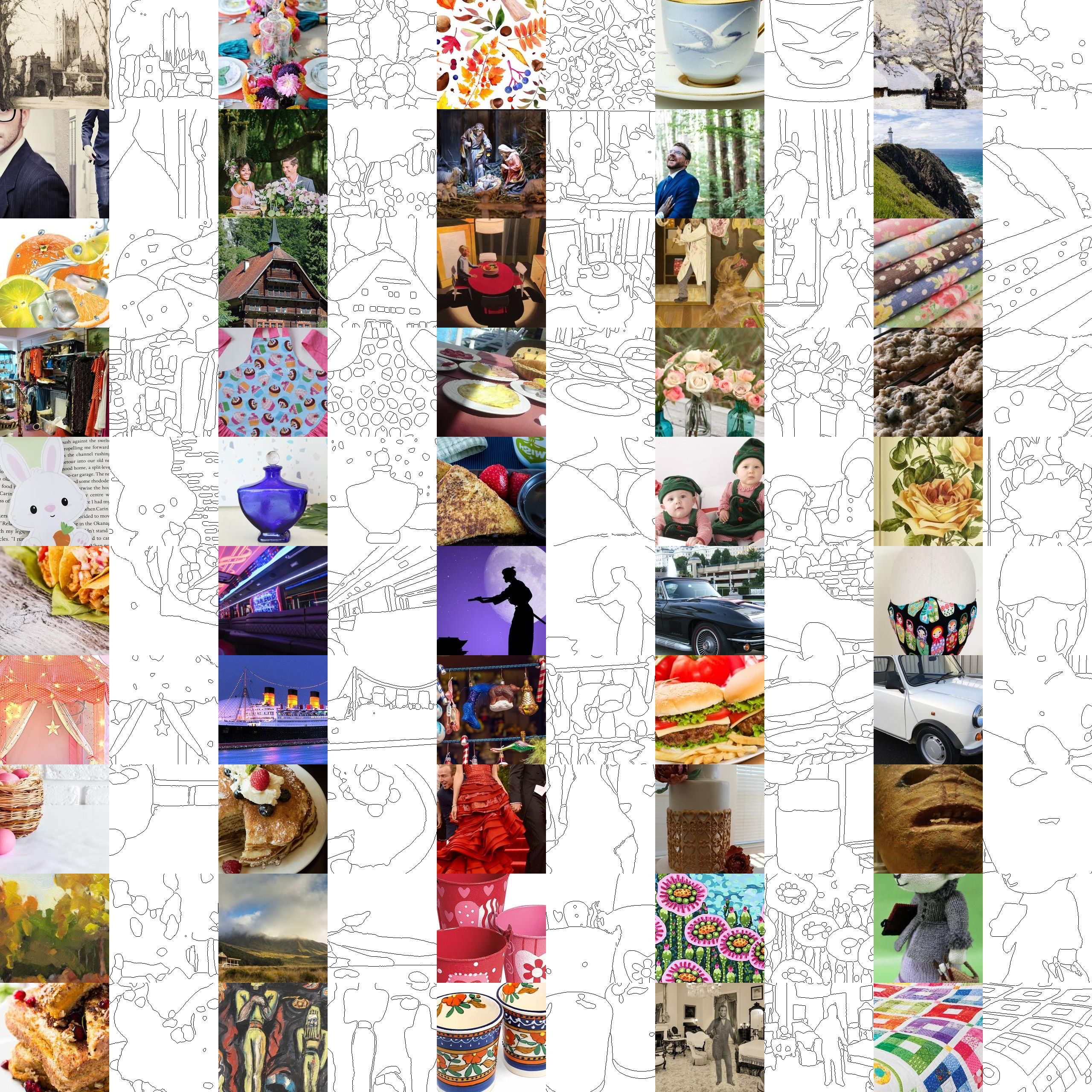}
    \caption{Example of synthetic edge dataset. Images are cropped to $256\times256$ for visualization.}
    \label{fig.dataset_preview_0}
\end{figure*}

\begin{figure*}
    \centering
    \includegraphics[width=\linewidth]{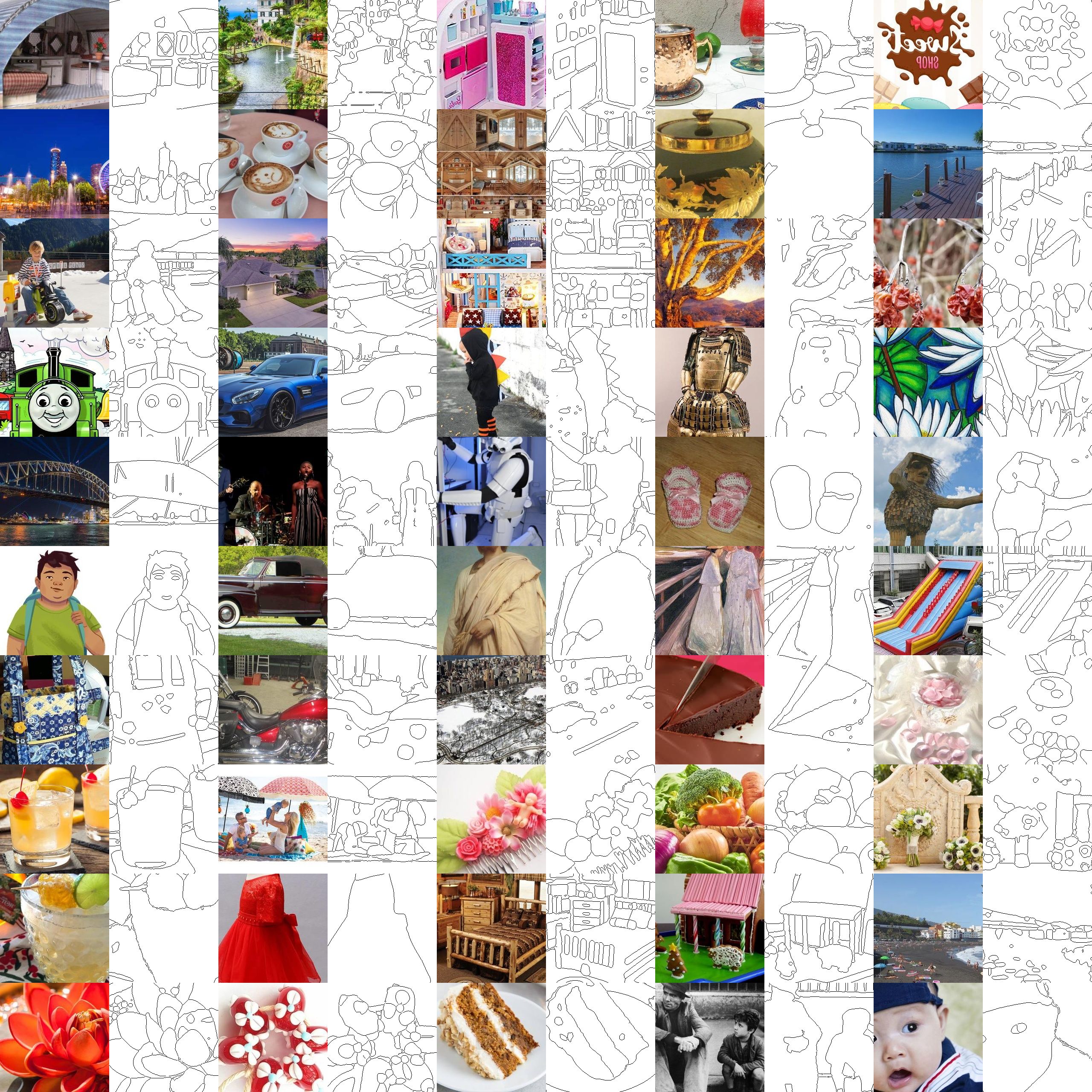}
    \caption{Example of synthetic edge dataset. Images are cropped to $256\times256$ for visualization.}
    \label{fig.dataset_preview_1}
\end{figure*}

\Cref{fig.dataset_preview_0,fig.dataset_preview_1} show random samples of generated synthetic dataset with source images from LAION~\cite{laion} and their edge side-by-side. Images are center cropped to $256\times256$ for better visualization, the actual source image and corresponding edge in the dataset has preserve their original aspect ratio.

\section{Related Works}

\noindent\textbf{Edge Detection} 
Deep learning-based edge detectors~\cite{hed,rcf,bdcn,edter,diffedge,uaed,muge} typically formulate edge detection as a pixel-wise binary classification problem optimized with a binary cross-entropy loss. 
HED~\cite{hed} introduces holistically-nested deep supervision with multi-scale side outputs to improve training stability and performance. 
RCF~\cite{rcf} aggregates richer convolutional features from multiple stages of the backbone to better capture both low-level and high-level cues. 
BDCN~\cite{bdcn} adopts a bi-directional cascade structure with layer-specific supervision to learn scale-aware edge responses. 
EDTER~\cite{edter} replaces purely convolutional backbones with a transformer-based architecture to exploit long-range dependencies for edge detection. 
DiffusionEdge~\cite{diffedge} formulates edge detection as a conditional diffusion process, using a diffusion probabilistic model to iteratively predict edge maps. 
UAED~\cite{uaed} explicitly models annotation uncertainty by leveraging multiple annotations to learn an uncertainty-aware edge predictor. 
MuGE~\cite{muge} extends this idea to multiple granularity levels, supervising edges at different granularity and strengths to better align with diverse annotation patterns. 
SAUGE~\cite{sauge} leverages the prior from segmentation models, training an adaption network to map the intermediate features of segmentation models to edge prediction.
However, as long as the task is treated as binary classification on thick binary edge maps and optimized with cross-entropy loss, the predictions tend to suffer from the thick-edge issue, where the output is a band of responses around the true contour rather than a single-pixel-wide edge as produced by human annotators.

\noindent\textbf{Crisp Edge Detection} 
To mitigate the thick-edge problem, several methods have been proposed to encourage crisp, thin edge predictions. 
CED~\cite{ced} adds side-refinement modules and supervision to sharpen boundary localization and suppress off-edge activations. 
LPCB~\cite{lpcb} formulates boundary prediction with a loss design and training strategy that emphasize precise, localized responses within each patch, thereby improving contour crispness. 
CATS~\cite{cats} analyzes and ``unmixes'' convolutional features, introducing feature unmixing and refinement modules to reduce blurry activations and produce sharper edges. 
Refined-label training~\cite{refined_label} observes that label noise and misaligned annotations are a key source of thick predictions, and therefore proposes a guided label refinement strategy to obtain cleaner, sharper edge labels for supervision. 
DiffusionEdge~\cite{diffedge}, by leveraging a diffusion generative backbone, also contributes to crisp edge detection by generating results that resembles to visually crisp edges.

\section{Limitations and Future Work}
\label{sec.limitation}
MEMO relies on recursive inference to refine edge predictions. While this iterative process is crucial for achieving high crispness, it also introduces non-trivial computational overhead, making the current model unsuitable for real-time edge detection on high-resolution images. 

There are several promising directions to improve the efficiency of MEMO, such as distilling the iterative refinement process into a single (or few) feed-forward passes, designing adaptive step schedules and early-stopping criteria that dynamically decide how many refinement steps are needed, or combining MEMO with lightweight backbones and multi-scale tiling strategies for large images. However, each of these directions would require substantial algorithmic and system-level engineering, and a careful study of the trade-off between accuracy and efficiency. We therefore leave them as important directions for future work.

\end{document}